\documentclass[10pt,twocolumn,letterpaper]{article}

\usepackage{cvpr}              

\usepackage[dvipsnames]{xcolor}
\usepackage{bbm}

\usepackage{multirow}
\usepackage{ctable}
\usepackage{comment}
\usepackage{makecell}
\usepackage{tabularx}

\usepackage{bbding}
\usepackage{pifont}
\usepackage{wasysym}
\usepackage{amssymb}
\usepackage[flushleft]{threeparttable}

\newcommand{\cmark}{\ding{51}}%
\newcommand{\xmark}{\ding{55}}%

\definecolor{cvprblue}{rgb}{0.21,0.49,0.74}
\usepackage[pagebackref,breaklinks,colorlinks,citecolor=cvprblue]{hyperref}

\def\paperID{15692} 
\def\confName{CVPR}
\def\confYear{2024}

\title{Versatile Medical Image Segmentation Learned from Multi-Source \\ Datasets via Model Self-Disambiguation}

\author{Xiaoyang Chen \and Hao Zheng \and Yuemeng Li \and Yuncong Ma \and Liang Ma \and Hongming Li \and Yong Fan \\
University of Pennsylvania\\
{\tt\small \{xiaoyang.chen,yong.fan\}@pennmedicine.upenn.edu}
}

\begin{document}
\maketitle
\begin{abstract}
A versatile medical image segmentation model applicable to images acquired with diverse equipment and protocols can facilitate model deployment and maintenance. However, building such a model typically demands a large, diverse, and fully annotated dataset, which is challenging to obtain due to the labor-intensive nature of data curation. To address this challenge, we propose a cost-effective alternative that harnesses multi-source data with only partial or sparse segmentation labels for training, substantially reducing the cost of developing a versatile model. We devise strategies for model self-disambiguation, prior knowledge incorporation, and imbalance mitigation to tackle challenges associated with inconsistently labeled multi-source data, including label ambiguity and modality, dataset, and class imbalances. Experimental results on a multi-modal dataset compiled from eight different sources for abdominal structure segmentation have demonstrated the effectiveness and superior performance of our method compared to state-of-the-art alternative approaches. We anticipate that its cost-saving features, which optimize the utilization of existing annotated data and reduce annotation efforts for new data, will have a significant impact in the field.
\end{abstract}
    
\section{Introduction}
\label{sec:intro}
Medical image segmentation plays a pivotal role in disease diagnosis \cite{de2018clinically,zhao2018deep,liu2021skullengine,babajide2022automated}, treatment planning \cite{tang2019clinically}, and biomedical research \cite{fischl2002whole,chen2023brain,li2021acenet,chen2023paired}. Models, tailored for specific applications, imaging modalities, and distinct anatomical regions, are ubiquitous and attract considerable attention \cite{wang2023dual}. Nonetheless, they often exhibit limited robustness and generalizability, largely due to insufficient training data. Additionally, the necessity to develop separate segmentation models for different organs and modalities poses scalability challenges, causing inefficient resource utilization and escalating development and maintenance costs.

Versatile image segmentation models show potential in overcoming the limitations of their specialized counterparts. However, their training typically requires a large, diverse, and fully annotated dataset, incurring high costs in data curation and annotation. As a result, only small-scale datasets are usually available, with annotations covering only a portion of anatomical structures or image slices, resulting in partial or sparse segmentation labels \cite{wang2024mixsegnet}. These datasets are typically curated by annotators focusing on labeling specific structures of interest while treating others as background. However, such selective annotation introduces ambiguity when interpreting unannotated regions, impeding the efficacy of image segmentation methods that rely on complete annotations. Specialized strategies are thus essential to effectively utilize datasets with ambiguous labels.

Multi-head segmentation models obviate the issue of labeling ambiguity by designing a distinct decoder for different datasets \cite{chen2019med3d,ji2023continual}. However, their inefficient memory usage hampers scalability. Dynamic models, such as the class-conditioned model \cite{dmitriev2019learning}, DoDNet \cite{zhang2021dodnet}, and CLIP-driven \cite{liu2023clip}, adeptly address partial labeling through conditional segmentation, enabling adjustments in their outputs for specific tasks. Nevertheless, dynamic models face their own challenges, such as training complexities, inefficiencies in inference due to multiple forward passes, and limitations in fully exploiting the benefits of fine-grained annotations, as class-specific parameters are optimized separately. Semi-supervised segmentation methods generate pseudo-labels for unannotated anatomical structures to facilitate conventional loss computation \cite{zhou2019prior,huang2020multi}. However, these methods require fully annotated data for initial fully supervised training, and the incorporation of inaccurate pseudo-labels in later training stage may degrade the model performance. Background modeling methods \cite{fang2020multi,shi2021marginal} dynamically compute losses for unannotated voxels to mitigate semantic drifts in partial annotations. Nevertheless, their requirement for fully annotated data limits their applicability in challenging scenarios. Notably, all these existing methods are unable to utilize sparsely labeled data (cf. Fig.~\ref{fig1}(c)).

In this study, we present a weakly-supervised approach for medical image segmentation, utilizing a large and diverse dataset with incomplete labeling from multiple sources. Our method utilizes a model self-disambiguation mechanism to tackle labeling ambiguity in both partially and sparsely annotated data. This is achieved by introducing two ambiguity-aware loss functions. Additionally, by leveraging prior knowledge of optimal predictions, we integrate a regularization term into the objective function. This helps reduce uncertainty in model predictions, particularly for challenging and unannotated voxels, thereby expediting convergence. To address imbalances in multi-source data, we propose a hierarchical sampling strategy. Our approach facilitates training a single versatile model using multi-source datasets and enables efficient inference in a single forward pass, predicting all anatomical structures simultaneously. Our contributions are three-fold:
\begin{itemize}
  \item We propose a weakly-supervised approach that leverages partially and sparsely labeled data to address data limitations in medical image segmentation. Remarkably, our approach exhibits impressive versatility and self-disambiguation capabilities, holding great promise for enhancing label efficiency and reducing the costs associated with model development, deployment, and maintenance.
  \item We employ hierarchical sampling to account for the imbalance issues in multi-source datasets and incorporate prior knowledge to improve the model performance.
  \item We showcase the proposed method's effectiveness on a multi-modality dataset of $2,960$ scans from eight distinct sources for abdominal organ segmentation. Our approach demonstrates substantially improved efficiency and effectiveness compared to state-of-the-art alternative methods.
\end{itemize}
\section{Related Work}
\label{sec:related}

\noindent \textbf{Category-specific models.} Developing separate models for different anatomical structures using annotated data from various sources is a straightforward strategy for leveraging multi-source datasets. However, this method is computationally complex and inefficient, as it requires training multiple models and processing test images through each model during inference. Additionally, it fails to capitalize on the benefits of fine-grained segmentation, which could improve feature representations and overall performance \cite{larsson2019fine}.

\noindent \textbf{Multi-head models.} Multi-head models \cite{chen2019med3d,ji2023continual} share an encoder but have separate decoders for each dataset. Yet, redundant structures in multiple decoders hinder scalability, and training decoders with limited and less diverse data may degrade model generalization.

\noindent \textbf{Dynamic models.} Dynamic models like the class-conditioned model \cite{dmitriev2019learning}, DoDNet \cite{zhang2021dodnet}, CLIP-driven model \cite{liu2023clip}, and Hermes \cite{gao2023training} utilize a unified model with task-adjustable outputs via a controller. However, they handle only one segmentation task at a time, causing inefficiencies during inference and limiting their exploitation of fine-grained segmentation benefits.

\noindent \textbf{Semi-supervised segmentation.} Semi-supervised segmentation methods \cite{zhou2019prior,huang2020multi} tackle partial labeling by generating pseudo-labels for unannotated anatomies, incorporating additional regularizations like anatomy size \cite{zhou2019prior} and inter-model consistency \cite{huang2020multi} to stabilize training. However, inaccuracies in pseudo-labels can impair model performance. Moreover, the necessity of a fully labeled dataset for pretraining \cite{zhou2019prior} and training multiple single-anatomy segmentation models \cite{huang2020multi} may limit practical applicability.

\noindent \textbf{Weakly-supervised segmentation.} Weakly-supervised segmentation methods utilize various forms of weak supervision, including image-level labels \cite{huang2018weakly}, bounding boxes \cite{kervadec2020bounding}, points \cite{en2022annotation,zhang2023weakly}, scribbles \cite{luo2022scribble,wang2024mixsegnet}, and incomplete annotations \cite{fang2020multi,shi2021marginal,bokhorst2018learning,nofallah2022segmenting}. Our work falls within this broad category, focusing on learning from ambiguous data labeled partially and sparsely. However, in contrast to existing methods that utilize image-level labels, bounding boxes, points and scribbles, none of which can attain comparable segmentation performance to voxel-wise supervision, or those tailored for training with partial labels, which struggle to leverage sparse labels, or those trained on data with sparse labels, requiring clear background definitions and annotations, our approach is designed to handle both partial and sparse labels, achieving highly competitive performance to fully supervised segmentation, even in the absence of background annotations.

\noindent \textbf{Background modeling.} Background modeling methods \cite{fang2020multi,shi2021marginal} address label ambiguity in partially labeled data by dynamically calculating the loss for unannotated voxels. These methods assume complete annotations of all identified organs within the volume and consolidate the predictions for all categories, excluding the annotated ones, into a distinct class during loss computation. Our approach bears similarities to these methods but offers enhanced capability in handling sparsely annotated data by relaxing labeling constraints. Notably, these methods still require fully annotated data during training, limiting their applicability in practical scenarios where such data is unavailable. In contrast, our method maintains effectiveness even when all training images are incompletely annotated.

\noindent \textbf{Segment anything model.} The ``segment anything'' model \cite{kirillov2023segment} and its variants, such as MedSAM \cite{ma2023segment} and SAM-Med2D \cite{cheng2023sam}, share the same goal as our work, aiming for a universal segmentation model applicable to various images and objects. However, these models presume the availability of a substantially large labeled dataset and do not attempt to handle practical challenges like label incompleteness and ambiguity. Moreover, these models are typically designed to generate segmentation results automatically without providing their semantic labels and are more suited for interactive use, requiring user input, such as a bounding box.

\section{Methods}

\begin{figure}[!t]
  \centering
  \includegraphics[width=0.44\textwidth]{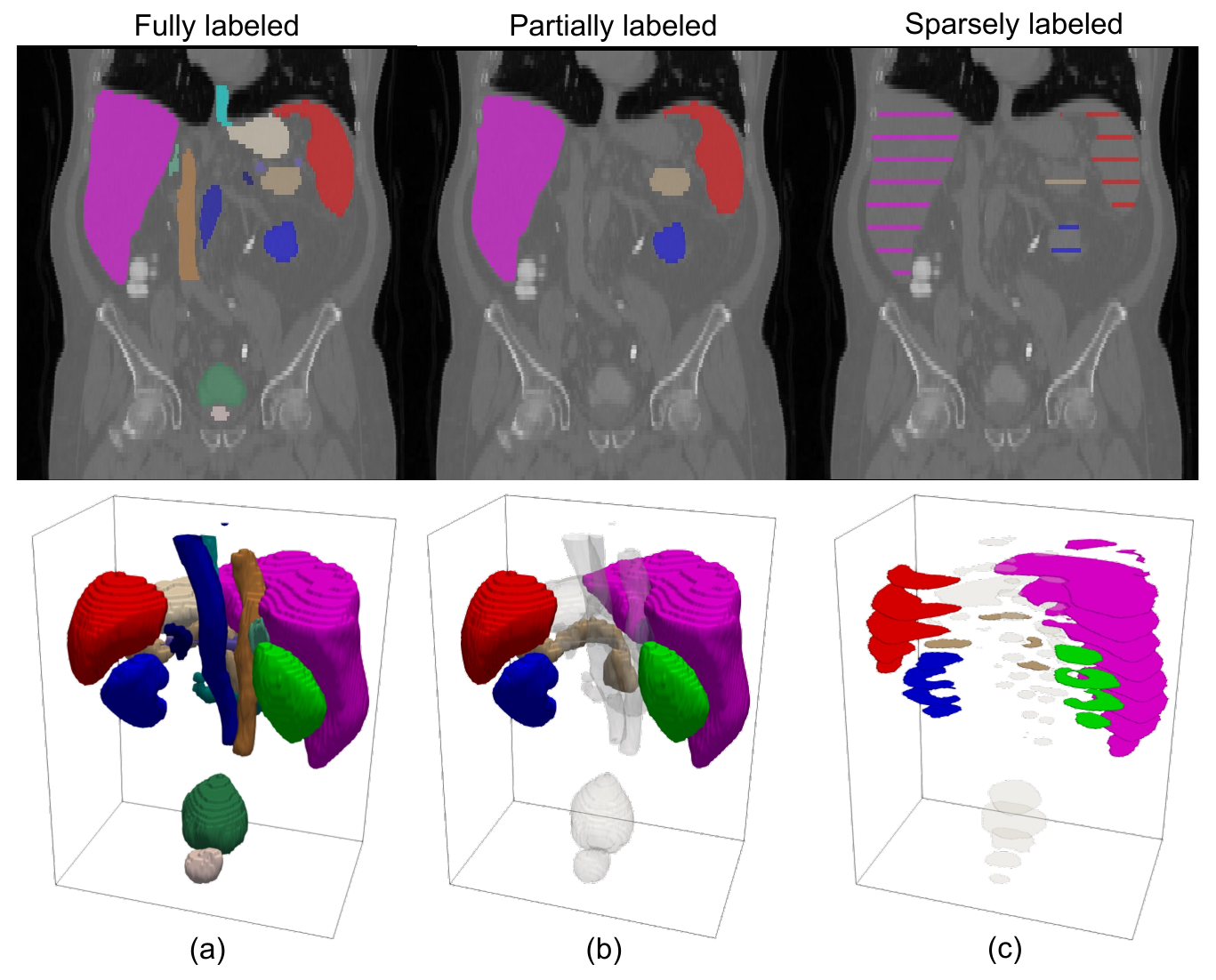}
  \caption{Illustrations of (a) fully labeled, (b) partially labeled, and (c) sparsely labeled images. The fully labeled image contains annotations for all anatomical structures of interest, the partially labeled image includes labels for a subset, and the sparsely labeled image provides annotations for only a fraction of the slices and structures. Note that annotated structures are fully marked within a particular volume (b) or slice (c).}
  \label{fig1}
\end{figure}

This section begins by introducing the motivation, objective and scope of our study. It then offers an overview of our proposed framework, followed by a detailed explanation of the key strategies employed in this research.
 
\subsection{Motivation, Objective \& Scope}
Given the challenges of obtaining a large, diverse, and fully annotated dataset for training versatile medical image segmentation models, as well as the increased accessibility and cost-effectiveness of weakly labeled data compared to the fully labeled data, our study pursues a cost-effective alternative utilizing two forms of weak supervision: partially labeled data and sparsely labeled data. Fig.~\ref{fig1} illustrates the differences between these data types, emphasizing variations in their annotation scopes and details.

It is important to clarify that the term ``sparsely labeled data'' here specifically refers to images with per-voxel annotations, rather than other types of weak annotations such as image-level labels, point annotations, scribbles, and bounding boxes, which are often categorized as sparse annotations in other studies. Nonetheless, our definition allows for flexibility: annotations for different slices and structures are independent, meaning that a structure annotated in one slice does not have to be annotated in other slices. Additionally, it is crucial to emphasize that sparsely labeled data encompasses a broader spectrum, with partially labeled data representing a specific category within it. By using these distinct terminologies, we highlight the differences between existing methods, primarily tailored for partially labeled data, and our approach, which accommodates a wider range of weakly labeled data. Our method excels at better data utilization and has the potential to enhance data accessibility.

\begin{figure*}[ht]
  \centering
  \includegraphics[width=0.88\textwidth]{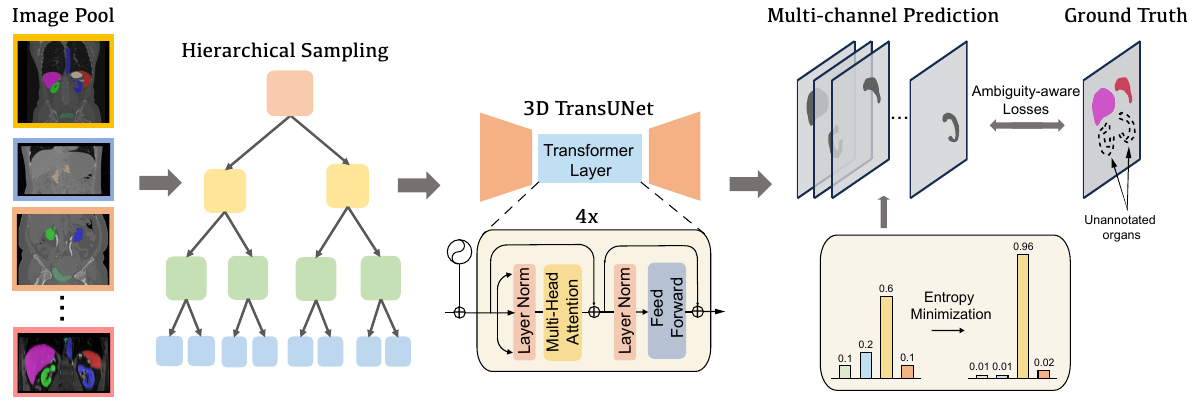}
  \caption{Overview of our approach. It trains a model by using hierarchical sampling for training example generation, 3D TransUNet as its base network, two ambiguity-aware losses and a prior knowledge-based entropy minimization regularization term for guidance.}
  \label{fig2}
\end{figure*}

\subsection{Overview}
As illustrated in Fig.~\ref{fig2}, our approach employs a hierarchical sampling technique to generate training examples from multi-source multi-modality datasets with ambiguous annotations. A 3D variant of TransUNet \cite{chen2021transunet} (3D TransUNet) is adopted as the base network for extracting per-voxel feature representations from the input. These representations are then processed by a segmentation head to produce multi-channel predictions. To address label ambiguity and ensure effective training, we integrate ambiguity-aware losses. Moreover, we incorporate prior knowledge to regularize the model training.

\subsection{Model Self-disambiguation}
When annotating medical images, it is a common practice for annotators to focus solely on labeling the anatomical structures of interest. Thus, a significant portion of voxels remains unannotated (with a default value of $0$) and is interpreted as background for each image. In a data collection comprising partially and/or sparsely labeled images from diverse sources, unannotated voxels in different images may contain various anatomical structures, leading to semantic ambiguity/drift within the background class. This semantic ambiguity presents significant challenges for fully supervised approaches due to conflicting supervision.

In this study, we tackle the challenges posed by semantic drift by computing the loss for unannotated voxels adaptively, considering both the possible categories for the unannotated voxels and the label type (\textit{i.e.}, partially labeled or sparsely labeled). Without loss of generality, let's consider a scenario where there are a total of $N$ anatomical structures of interest, denoted by $\Phi_{N}$, and each training example has only a fraction of its slices annotated. In each annotated slice, the annotation may cover only a subset of $M$ out of $N$ structures, where $1 \leq M \leq N$. Generally, this subset can comprise any combination and be denoted as $\Phi_{M}$ $=$ $\{i_{1}, i_{2}, \dots, i_{M}\}$, where $1 \leq i_{p} < i_{q} \leq N$ for any $p < q$. For the annotated voxels in a given slice, their labels are definitive and offer clear supervision. However, the true labels for the unannotated voxels in the same slice are unknown. It is only certain that these voxels may belong to either the ``real'' background category or any class in $\Phi_{N}{\setminus}\Phi_{M}$, representing the difference between sets $\Phi_{N}$ and $\Phi_{M}$, \textit{i.e.}, $\{x \mid x \in \Phi_{N} \text{ and } x \notin \Phi_{M}\}$. Therefore, we adaptively adjust the loss calculation for the unannotated voxels to accommodate label ambiguity. We adopt the following ambiguity-aware focal cross-entropy loss ($\mathcal{L}_{\text{focal\_ce}}$) and dice loss ($\mathcal{L}_{\text{dice}}$), calculated slice-wise, as the objective:

\begin{equation}
\mathcal{L}_{\text{focal\_ce}} = \frac{1}{N_{v}} \sum_{c \in \{0\} \cup \Phi_{M}}\sum_{i=1}^{N_{v}} \mathbbm{1}_{y_{i}=c} (1-\tilde{p}_{ic})^2\log \tilde{p}_{ic},
\end{equation}
\begin{equation}
\mathcal{L}_{\text{dice}}=1-\frac{1}{|\Phi_{M}|+1}  \sum_{c \in \{0\} \cup \Phi_{M}} \frac{2 \cdot \text{TP}_{c} + \epsilon}{2 \cdot \text{TP}_{c} + \text{FP}_{c} + \text{FN}_{c} + \epsilon},
\end{equation}
where $\mathbbm{1}$ denotes an indicator function, $|\cdot|$ is the cardinality, $N_{v}$ represents the number of pixels in the slice, $\text{TP}_{c} = \sum_{i=1}^{N_{v}} \tilde{p}_{ic} \tilde{y}_{ic}$, $\text{FP}_{c} = \sum_{i=1}^{N_{v}} \tilde{p}_{ic} (1-\tilde{y}_{ic})$ and $\text{FN}_{c} = \sum_{i=1}^{N_{v}} (1-\tilde{p}_{ic}) \tilde{y}_{ic}$ are the soft values for the true positive, false positive and false negative respectively, $\epsilon$ is set to $1$ to avoid division by 0,
\begin{equation}
\tilde{p}_{ic}=
\begin{cases}
p_{ic}, &c \in \Phi_{M},\\
\sum_{j \not\in \Phi_{M}} p_{ij}, &c = 0,
\end{cases}
\end{equation}
\begin{equation}
\tilde{y}_{ic}=
\begin{cases}
y_{ic}, &c \in \Phi_{M},\\
\sum_{j \not\in \Phi_{M}} y_{ij}, &c = 0,
\end{cases}
\end{equation}
$p_{ic}$ and $y_{ic}$ represent the $c$-th element of the probability vector and the one-hot encoded vector for the expert label for the $i$-th pixel, respectively.

Unlike methods that simply treat unannotated voxels as background, potentially misleading the model, or those that overlook voxels with ambiguous labels in the loss calculation, leading to incorrect predictions for voxels not belonging to any specific structure, our ambiguity-aware losses enable the model to self-disambiguate during training and infer the correct labels for all voxels.

It is noteworthy that for partially labeled data, the loss computation can be simplified. The ambiguity-aware losses can be calculated for each volume rather than slicewise, with $N_{v}$ representing the number of voxels, and $\Phi_{M}$ denoting the set of annotated structures within the volume.

\subsection{Prior Knowledge Incorporation}
We have the prior knowledge that each voxel/pixel corresponds to a single label in multi-class medical image segmentation tasks. When confronted with challenging and unannotated voxels, the model encounters difficulty in determining their classes, leading to high-entropy predictions. Our hypothesis is that reducing this uncertainty can improve the differentiation between categories and accelerate a more reliable convergence during the optimization process. We thus incorporate this prior knowledge into the training process, encouraging the model to produce more confident and informative predictions. This is achieved by regularizing the model training to minimize the Shannon entropy below.
\begin{equation}
\mathcal{L}_{\text{reg}} = -\frac{1}{N_{v}} \sum_{i=1}^{N_{v}} \sum_{c=0}^{N} p_{ic} \log p_{ic},
\end{equation}
Notably, this regularizer is class-agnostic and can be applied to both annotated and unannotated voxels.

\subsection{Imbalance Mitigation}
Medical image segmentation models often encounter challenges in maintaining consistent performance across diverse domains due to variations in imaging modalities, equipment, imaging protocols, and patient demographics. While aggregating data from multiple sources can enrich training data diversity and bolster model robustness, it may also introduce imbalances at the modality, dataset, and class levels. Neglecting these issues during model training could result in inferior performance, particularly on underrepresented modalities, datasets, and categories.

However, such imbalance issues have not been well addressed in existing methods that utilize multi-source partially labeled data \cite{zhou2019prior,huang2020multi,fang2020multi,shi2021marginal}. To tackle these challenges, we propose a hierarchical sampling approach. During training, we initiate the sampling by selecting images based on the type of anatomical structure, thereby narrowing down the number of eligible images. The chosen structure determines the location of the training image patch center. Subsequently, we conduct a secondary sampling based on the modality of the medical images within the subset of images from the first level, enabling us to focus on images that belong to the chosen modality. Next, we draw a sample from the candidate pool based on the dataset of origin for each image, ensuring equitable treatment for images from various sources. Finally, we select an image from the chosen dataset for training. The proposed strategy enables us to account for the variations across domains, ultimately ensuring a balanced representation of the training data.

\subsection{Overall Objective}
 The overall objective for training ($\mathcal{L}$) is a weighted combination of the uncertainty-aware focal cross-entropy loss ($\mathcal{L}_{\text{focal\_ce}}$), uncertainty-aware dice loss ($\mathcal{L}_{\text{dice}}$), and the Shannon entropy minimization regularization term ($\mathcal{L}_{\text{reg}}$):
\begin{equation}
\mathcal{L} = \mathcal{L}_{\text{focal\_ce}} + \mathcal{L}_{\text{dice}} + \lambda\mathcal{L}_{\text{reg}}.
\end{equation}
where the hyper-parameter $\lambda$ is set to $3$ for training examples without annotations to mitigate the null effects of $\mathcal{L}_{\text{focal\_ce}}$ and $\mathcal{L}_{\text{dice}}$, and $1$ otherwise.
\section{Experiments and Results}

\begin{figure*}[!h]
    \begin{subfigure}{0.4\textwidth}
        \centering
        \vspace{0.16cm}
        \includegraphics[width=\textwidth]{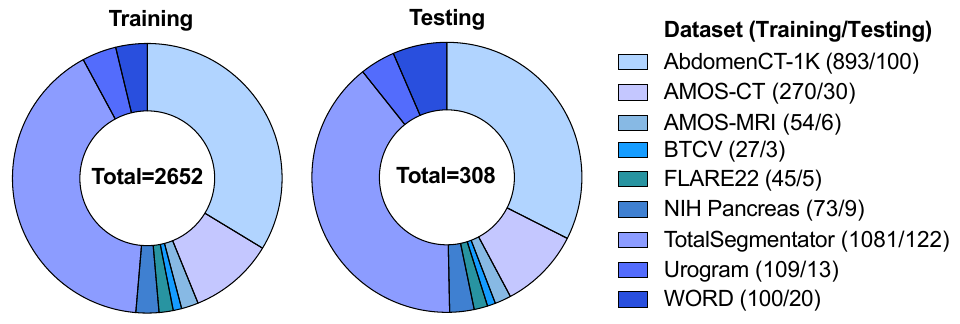}
        \caption{}
    \end{subfigure}
    \hfill
    \begin{subfigure}{0.6\textwidth}
        \centering
        \resizebox{0.93\textwidth}{!}{%
        \begin{tabular}{c|c|c|c|c|c|c|c|c|c|c|c|c|c|c|c|c}
        \toprule
        \textbf{Dataset} & \textbf{Sp} & \textbf{RK} & \textbf{LK} & \textbf{GB} & \textbf{Eso} & \textbf{L} & \textbf{St} & \textbf{A} & \textbf{PC} & \textbf{Pan} & \textbf{RAG} & \textbf{LAG} & \textbf{Duo} & \textbf{B} & \textbf{PU} & \textbf{PSV} \\
        \midrule
        AbCT-1K  & \cmark & \cmark & \cmark & \xmark & \xmark & \cmark & \xmark & \xmark & \xmark & \cmark & \xmark & \xmark & \xmark & \xmark & \xmark & \xmark \\
        AMOS-CT & \cmark & \cmark & \cmark & \cmark & \cmark & \cmark & \cmark & \cmark & \cmark & \cmark & \cmark & \cmark & \cmark & \cmark & \cmark & \xmark \\
        AMOS-MRI & \cmark & \cmark & \cmark & \cmark & \cmark & \cmark & \cmark & \cmark & \cmark & \cmark & \cmark & \cmark & \cmark & \cmark & \cmark & \xmark \\
        BTCV & \cmark & \cmark & \cmark & \cmark & \cmark & \cmark & \cmark & \cmark & \cmark & \cmark & \cmark & \cmark & \xmark & \xmark & \xmark & \cmark \\
        FLARE22 & \cmark & \cmark & \cmark & \cmark & \cmark & \cmark & \cmark & \cmark & \cmark & \cmark & \cmark & \cmark & \cmark & \xmark & \xmark & \xmark \\
        NIH-Pan & \xmark & \xmark & \xmark & \xmark & \xmark & \xmark & \xmark & \xmark & \xmark & \cmark & \xmark & \xmark & \xmark & \xmark & \xmark & \xmark \\
        TotalSeg & \cmark & \cmark & \cmark & \cmark & \cmark & \cmark & \cmark & \cmark & \cmark & \cmark & \cmark & \cmark & \cmark & \cmark & \xmark & \cmark \\
        Urogram & \xmark & \cmark & \cmark & \xmark & \xmark & \xmark & \xmark & \xmark & \xmark & \xmark & \xmark & \xmark & \xmark & \cmark & \xmark & \xmark \\
        WORD & \cmark & \cmark & \cmark & \cmark & \cmark & \cmark & \cmark & \xmark & \xmark & \cmark & \cmark & \cmark & \cmark & \cmark & \xmark & \xmark \\
        \bottomrule
        \end{tabular}}
        \caption{}
    \end{subfigure}
    \caption{(a): Training and testing image composition. (b): Annotated anatomical structures in different datasets.}
    \label{fig3}
\end{figure*}

\subsection{Experiment Setup}

\noindent \textbf{Dataset.} We curated a dataset of $2,960$ volumetric images from eight sources, including seven public datasets (AbdomenCT-1K (AbCT-1K) \cite{ma2021abdomenct}, AMOS \cite{ji2022amos}, BTCV \cite{landman2015miccai}, FLARE 2022 (FLARE22) \cite{ma2023unleashing}, NIH pancreas (NIH-Pan)\cite{roth2015deeporgan}, TotalSegmentator (TotalSeg) \cite{wasserthal2022totalsegmentator}, and WORD \cite{luo2021word}) and one private dataset (Urogram). Notably, the AMOS dataset is multi-modality, featuring $300$ labeled CT and $60$ labeled MRI images, which we segregated into two subsets, AMOS-CT and AMOS-MRI. We demonstrated the model's self-disambiguation capability by segmenting $16$ abdominal structures, namely spleen (Sp), right kidney (RK), left kidney (LK), gallbladder (GB), esophagus (Eso), liver (L), stomach (St), aorta (A), postcava (PC), pancreas (Pan), right adrenal gland (RAG), left adrenal gland (LAG), duodenum (Duo), bladder (B), prostate/uterus (PU), and portal vein and splenic vein (PSV). Although TotalSeg and WORD provided masks for anatomical structures beyond our scope, we retained only the pertinent ones. In WORD, RK and RAG are assigned with the same labels as LK and LAG, respectively. We thus partitioned segments for kidney and adrenal gland into connected components and automatically assigned labels to the largest two, assisted by a model trained without WORD. The same was applied to AbCT-1K to separate RK and LK. For cases with only one kidney or adrenal gland, we manually verified and assigned the correct labels. Approximately $90\%$ of the images were selected for training, with the remainder reserved for evaluation purposes. Details about each dataset are outlined in Fig.~\ref{fig3}. Note that we have removed corrupted images, and \textbf{no} images used in this study are fully annotated.

\noindent \textbf{Data preprocessing.} To facilitate hierarchical sampling, we added a prefix to each image name indicating the dataset it comes from and its modality. For example, a CT image from AMOS initially named ``amos\_0001.nii.gz'' was renamed ``amos\_ct\_amos\_0001.nii.gz'' after prefixing. Additionally, we standardized all images to lie in a common coordinate system to ease model training with images in varied orientations. All data were resampled to a uniform spacing of $2\times2\times2$ $\text{mm}^{3}$. Intensity values in CT images were clipped at $-400$ and $400$ HU, while for MRI images, the clipping was done at the 1st and 99th percentiles of the intensity distribution. Finally, the intensity values were normalized to the range of $[0, 1]$.

\begin{table}[t]
        \centering
        \caption{Method performance comparison.}
        \resizebox{0.7\columnwidth}{!}{%
        \begin{tabular}{c|c|c}
            \toprule
            \textbf{Method} & \textbf{Base} & \textbf{DSC} [\%] \\
            \midrule
            DoDNet \cite{zhang2021dodnet} & 3D TransUNet & $83.5 \scriptstyle{\pm} 15.9$ \\
            CLIP-driven \cite{liu2023clip} & 3D TransUNet & $83.3 \scriptstyle{\pm} 16.3$ \\
            \midrule
            \multirow{4}{*}{\centering Ours} & Unet++ \cite{zhou2019unet++}  & $87.4 \scriptstyle{\pm} 8.5$ \\
             & Swin UNETR \cite{hatamizadeh2021swin} & $86.9 \scriptstyle{\pm} 8.5$ \\
             & MedNeXt \cite{roy2023mednext} & $88.4 \scriptstyle{\pm} 7.3$ \\
             & 3D TransUNet & $\textbf{88.7} \scriptstyle{\pm} \textbf{7.0}$ \\
            \bottomrule
         \end{tabular}}
        \label{tab2}
\end{table}

\noindent \textbf{Implementation details.} PyTorch was used to implement the proposed method. We employed the AdamW optimizer \cite{loshchilov2017decoupled} with an initial learning rate of $0.001$ and a polynomial learning rate scheduler with a decay of $0.9$. Data augmentations, such as random rotation and scaling, were applied during the training process. Unless otherwise specified, the default patch size and number of iterations were set to $112\times112\times112$ and $200,000$, respectively. Distributed data parallel was used to enhance training efficiency. All experiments were conducted on a single node with $8$ NVIDIA Titan Xp GPUs. The effective batch size was $8$ for all experiments, and all models were trained from scratch.

\noindent \textbf{Evaluation metric.} The Dice Similarity Coefficient (DSC, $\%$) was used for performance evaluation. Since annotations were limited to a subset of anatomical structures in each image, and the region of interest (ROI) varied across images, only those annotated structures within the ROI were included for quantitative evaluation. Notably, the segmentation difficulty varied across different structures, and the number of annotated structures differed among datasets, leading to significant variation in the quantitative values across datasets. It is noteworthy that all the reported performances were obtained with a single model, not through ensemble learning techniques.
 
\begin{table*}[htbp]
    \hspace{0.25cm}
    \begin{minipage}{0.64\textwidth}
        \centering
        \captionsetup{font=scriptsize}
        \caption{Comparison of overall and PU segmentation performance (DSC, \%) with varying numbers of datasets used for training. The ``3 Sets'' experiment exclusively involves AMOS, BTCV, and FLARE22 datasets.}
        \resizebox{0.32\textwidth}{!}{%
        \begin{tabular}{c|c|c}
        \toprule
        \textbf{Setting} & \textbf{Overall} & \textbf{PU} \\
        \midrule
        3 Sets & $83.5  \scriptstyle{\pm} 13.9$ & $75.8 \scriptstyle{\pm} 21.6$ \\
        8 Sets & $\textbf{88.7} \scriptstyle{\pm} \textbf{7.0}$ & $\textbf{79.2} \scriptstyle{\pm} \textbf{17.6}$ \\
        \bottomrule
        \end{tabular}}
        \label{tab8}
    \end{minipage}
    \hspace{-0.2cm}
    \begin{minipage}{0.36\textwidth}
        \centering
        \captionsetup{font=scriptsize}
        \caption{Performance with different sampling methods.}
        \resizebox{0.56\textwidth}{!}{%
        \begin{tabular}{c|c}
        \toprule
        \textbf{Sampling Approach} & \textbf{DSC} [\%] \\
        \midrule
        Random & $87.5 \scriptstyle{\pm} 7.7$ \\
        MDC & $87.9 \scriptstyle{\pm} 7.7$ \\
        CMD (Default) & $\textbf{88.7} \scriptstyle{\pm} \textbf{7.0}$ \\
        \bottomrule
        \end{tabular}}
        \label{tab6}
    \end{minipage}
\end{table*}

\begin{table*}[!h]
\centering
\caption{Performance (DSC, \%) comparison among DoDNet, CLIP-driven, and the proposed method on each anatomical structure.}\label{tab3}
\resizebox{0.9\textwidth}{!}{%
\begin{tabular}{c|c|c|c|c|c|c|c|c|c|c|c|c|c|c|c|c|c|c}
\toprule
\textbf{Method} & \textbf{Base} & \textbf{Sp} & \textbf{RK} & \textbf{LK} & \textbf{GB} & \textbf{Eso} & \textbf{L} & \textbf{St} & \textbf{A} & \textbf{PC} & \textbf{Pan} & \textbf{RAG} & \textbf{LAG} & \textbf{Duo} & \textbf{B} & \textbf{PU} & \textbf{PSV} & \textbf{Average} \\
\midrule
\multirow{2}{*}{\centering DoDNet}  & \multirow{2}{*}{3D TransUNet} & $92.6$ & $90.6$ & $89.8$ & $73.3$ & $76.6$ & $93.2$ & $85.8$ & $84.8$ & $82.5$ & $81.6$ & $69.6$ & $72.3$ & $69.6$ & $83.6$ & $59.2$ & $75.7$ & $80.0$ \\
                                        & & $\scriptstyle{\pm} 11.9$ & $\scriptstyle{\pm} 13.2$ & $\scriptstyle{\pm} 13.9$ & $\scriptstyle{\pm} 28.3$ & $\scriptstyle{\pm} 14.6$ & $\scriptstyle{\pm} 15.1$ & $\scriptstyle{\pm} 19.8$ & $\scriptstyle{\pm} 22.5$ & $\scriptstyle{\pm} 19.7$ & $\scriptstyle{\pm} 14.3$ & $\scriptstyle{\pm} 18.9$ & $\scriptstyle{\pm} 17.9$ & $\scriptstyle{\pm} 21.3$ & $\scriptstyle{\pm} 17.3$ & $\scriptstyle{\pm} 29.5$ & $\scriptstyle{\pm} 20.5$ & $\scriptstyle{\pm} 18.7$ \\
\multirow{2}{*}{\centering CLIP-driven} & \multirow{2}{*}{3D TransUNet} & $92.5$ & $89.0$ & $90.5$ & $74.5$ & $76.8$ & $93.1$ & $86.2$ & $84.4$ & $81.7$ & $81.6$ & $70.7$ & $73.4$ & $69.3$ & $85.9$ & $68.2$ & $74.5$ & $80.7$ \\
                                       & & $\scriptstyle{\pm} 11.8$ & $\scriptstyle{\pm} 16.6$ & $\scriptstyle{\pm} 13.0$ & $\scriptstyle{\pm} 27.7$ & $\scriptstyle{\pm} 14.6$ & $\scriptstyle{\pm} 15.2$ & $\scriptstyle{\pm} 20.0$ & $\scriptstyle{\pm} 21.9$ & $\scriptstyle{\pm} 21.0$ & $\scriptstyle{\pm} 14.1$ & $\scriptstyle{\pm} 17.7$ & $\scriptstyle{\pm} 17.4$ & $\scriptstyle{\pm} 22.7$ & $\scriptstyle{\pm} 16.5$ & $\scriptstyle{\pm} 30.3$ & $\scriptstyle{\pm} 21.7$ & $\scriptstyle{\pm} 18.9$ \\
\midrule
\multirow{8}{*}{\centering Ours} & \multirow{2}{*}{\centering UNet++} & $94.9$ & $93.1$ & $93.5$ & $78.2$ & $81.5$ & $96.2$ & $90.3$ & $90.7$ & $85.4$ & $84.1$ & $74.1$ & $75.2$ & $73.9$ & $88.2$ & $76.0$ & $75.8$ & $84.5$ \\
                                  & & $\scriptstyle{\pm} 6.9$ & $\scriptstyle{\pm} 9.0$ & $\scriptstyle{\pm} 4.9$ & $\scriptstyle{\pm} 23.2$ & $\scriptstyle{\pm} 8.6$ & $\scriptstyle{\pm} 6.8$ & $\scriptstyle{\pm} 13.0$ & $\scriptstyle{\pm} 11.2$ & $\scriptstyle{\pm} 12.7$ & $\scriptstyle{\pm} 9.4$ & $\scriptstyle{\pm} 12.5$ & $\scriptstyle{\pm} \textbf{13.3}$ & $\scriptstyle{\pm} 17.9$ & $\scriptstyle{\pm} 15.0$ & $\scriptstyle{\pm} 23.6$ & $\scriptstyle{\pm} 17.0$ & $\scriptstyle{\pm} 12.8$ \\
 & \multirow{2}{*}{\centering Swin UNETR} & $94.7$ & $92.8$ & $93.0$ & $76.4$ & $80.0$ & $96.1$ & $90.0$ & $90.4$ & $85.4$ & $83.0$ & $73.1$ & $74.0$ & $72.6$ & $88.4$ & $74.8$ & $74.8$ & $83.7$ \\
                                  & & $\scriptstyle{\pm} 7.3$ & $\scriptstyle{\pm} 9.7$ & $\scriptstyle{\pm} 6.4$ & $\scriptstyle{\pm} \textbf{22.9}$ & $\scriptstyle{\pm} 9.0$ & $\scriptstyle{\pm} 6.6$ & $\scriptstyle{\pm} 12.5$ & $\scriptstyle{\pm} 10.1$ & $\scriptstyle{\pm} 11.6$ & $\scriptstyle{\pm} 9.9$ & $\scriptstyle{\pm} 13.1$ & $\scriptstyle{\pm} 14.9$ & $\scriptstyle{\pm} 17.1$ & $\scriptstyle{\pm} 14.3$ & $\scriptstyle{\pm} 22.2$ & $\scriptstyle{\pm} 16.1$ & $\scriptstyle{\pm} 12.7$ \\
 & \multirow{2}{*}{\centering MedNeXt} & $95.0$ & $93.3$ & $93.7$ & $\textbf{78.4}$ & $\textbf{83.2}$ & $\textbf{96.5}$ & $91.3$ & $91.9$ & $86.7$ & $\textbf{85.1}$ & $\textbf{75.5}$ & $76.4$ & $76.3$ & $88.9$ & $77.5$ & $\textbf{77.6}$ & $84.5$ \\
                                  & & $\scriptstyle{\pm} 8.9$ & $\scriptstyle{\pm} 9.5$ & $\scriptstyle{\pm} 6.9$ & $\scriptstyle{\pm} 23.9$ & $\scriptstyle{\pm} \textbf{7.5}$ & $\scriptstyle{\pm} \textbf{6.4}$ & $\scriptstyle{\pm} \textbf{12.1}$ & $\scriptstyle{\pm} 8.3$ & $\scriptstyle{\pm} 12.3$ & $\scriptstyle{\pm} \textbf{8.9}$ & $\scriptstyle{\pm} \textbf{11.3}$ & $\scriptstyle{\pm} 13.4$ & $\scriptstyle{\pm} \textbf{15.9}$ & $\scriptstyle{\pm} 14.2$ & $\scriptstyle{\pm} 23.7$ & $\scriptstyle{\pm} \textbf{15.5}$ & $\scriptstyle{\pm} 12.4$ \\
 & \multirow{2}{*}{\centering 3D TransUNet} & $\textbf{95.3}$ & $\textbf{93.8}$ & $\textbf{94.0}$ & $77.7$ & $82.4$ & $\textbf{96.5}$ & $\textbf{91.5}$ & $\textbf{92.5}$ & $\textbf{87.0}$ & $\textbf{85.1}$ & $75.4$ & $\textbf{76.5}$ & $\textbf{76.4}$ & $\textbf{90.3}$ & $\textbf{79.2}$ & $\textbf{77.6}$ & $\textbf{85.7}$ \\
                                  & & $\scriptstyle{\pm} \textbf{6.6}$ & $\scriptstyle{\pm} \textbf{7.3}$ & $\scriptstyle{\pm} \textbf{5.2}$ & $\scriptstyle{\pm} 25.0$ & $\scriptstyle{\pm} 9.0$ & $\scriptstyle{\pm} \textbf{6.4}$ & $\scriptstyle{\pm} 12.3$ & $\scriptstyle{\pm} \textbf{6.5}$ & $\scriptstyle{\pm} \textbf{11.2}$ & $\scriptstyle{\pm} 9.3$ & $\scriptstyle{\pm} 12.0$ & $\scriptstyle{\pm} 14.6$ & $\scriptstyle{\pm} 16.6$ & $\scriptstyle{\pm} \textbf{13.3}$ & $\scriptstyle{\pm} \textbf{17.6}$ & $\scriptstyle{\pm} 16.5$ & $\scriptstyle{\pm} \textbf{11.9}$ \\                                                                    
\bottomrule
\end{tabular}}
\end{table*}

\begin{table*}[htbp]
        \centering
        \caption{Performance (DSC, \%) comparison among DoDNet, CLIP-driven and the proposed method on each dataset.}
        \resizebox{0.78\textwidth}{!}{%
        \begin{tabular}{c|c|c|c|c|c|c|c|c|c|c}
        \toprule
        \textbf{Method} & \textbf{Base} & \textbf{AbCT-1K} & \textbf{AMOS-CT} & \textbf{AMOS-MRI} & \textbf{BTCV} & \textbf{FLARE22} & \textbf{NIH-Pan} & \textbf{TotalSeg} & \textbf{Urogram} & \textbf{WORD} \\
        \midrule
        \multirow{2}{*}{DoDNet} & \multirow{2}{*}{3D TransUNet} & $91.3$ & $81.5$ & $80.5$ & $76.0$ & $89.0$ & $82.4$ & $77.6$ & $90.9$ & $79.3$ \\
                                       & & $\scriptstyle{\pm} 2.9$ & $\scriptstyle{\pm} 9.0$ & $\scriptstyle{\pm} 9.5$ & $\scriptstyle{\pm} 5.4$ & $\scriptstyle{\pm} 1.5$ & $\scriptstyle{\pm} 4.5$ & $\scriptstyle{\pm} 22.5$ & $\scriptstyle{\pm} 2.7$ & $\scriptstyle{\pm} 4.9$ \\
        \multirow{2}{*}{CLIP-driven} & \multirow{2}{*}{3D TransUNet} & $91.3$ & $82.0$ & $80.8$ & $76.6$ & $89.3$ & $81.9$ & $76.8$ & $90.3$ & $80.6$ \\
                                       & & $\scriptstyle{\pm} 3.2$ & $\scriptstyle{\pm} 9.3$ & $\scriptstyle{\pm} 9.3$ & $\scriptstyle{\pm} 5.3$ & $\scriptstyle{\pm} 1.2$ & $\scriptstyle{\pm} 4.0$ & $\scriptstyle{\pm} 23.1$ & $\scriptstyle{\pm} 3.7$ & $\scriptstyle{\pm} 4.2$ \\
        \midrule
        \multirow{8}{*}{Ours} & \multirow{2}{*}{UNet++} & $92.9$ & $84.9$ & $\textbf{81.6}$ & $\textbf{79.4}$ & $90.5$ & $83.6$ & $84.2$ & $93.1$ & $83.5$ \\
                                       & & $\scriptstyle{\pm} 2.3$ & $\scriptstyle{\pm} 6.9$ & $\scriptstyle{\pm} 8.8$ & $\scriptstyle{\pm} 5.9$ & $\scriptstyle{\pm} 0.9$ & $\scriptstyle{\pm} 5.2$ & $\scriptstyle{\pm} 10.5$ & $\scriptstyle{\pm} 1.8$ & $\scriptstyle{\pm} 4.5$ \\
         & \multirow{2}{*}{Swin UNETR} & $92.8$ & $83.6$ & $80.6$ & $78.8$ & $90.2$ & $82.7$ & $83.4$ & $93.1$ & $83.1$ \\
                                       & & $\scriptstyle{\pm} 2.0$ & $\scriptstyle{\pm} 7.9$ & $\scriptstyle{\pm} 6.5$ & $\scriptstyle{\pm} 5.9$ & $\scriptstyle{\pm} 1.3$ & $\scriptstyle{\pm} 5.2$ & $\scriptstyle{\pm} 10.1$ & $\scriptstyle{\pm} \textbf{1.5}$ & $\scriptstyle{\pm} 4.3$ \\
         & \multirow{2}{*}{MedNeXt} & $93.3$ & $\textbf{85.6}$ & $\textbf{81.6}$ & $78.8$ & $90.8$ & $84.2$ & $85.9$ & $\textbf{93.8}$ & $\textbf{84.2}$ \\
                                       & & $\scriptstyle{\pm} 2.4$ & $\scriptstyle{\pm} 7.2$ & $\scriptstyle{\pm} \textbf{8.1}$ & $\scriptstyle{\pm} 6.7$ & $\scriptstyle{\pm} \textbf{0.6}$ & $\scriptstyle{\pm} 4.2$ & $\scriptstyle{\pm} 8.5$ & $\scriptstyle{\pm} \textbf{1.5}$ & $\scriptstyle{\pm} 3.9$ \\
         &  \multirow{2}{*}{3D TransUNet} & $\textbf{93.5}$ & $85.5$ & $\textbf{80.5}$ & $78.7$ & $\textbf{90.9}$ & $\textbf{84.7}$ & $\textbf{86.4}$ & $93.6$ & $84.1$ \\
                                       & & $\scriptstyle{\pm} \textbf{1.9}$ & $\scriptstyle{\pm} \textbf{6.6}$ & $\scriptstyle{\pm} 9.3$ & $\scriptstyle{\pm} \textbf{5.6}$ & $\scriptstyle{\pm} 0.7$ & $\scriptstyle{\pm} \textbf{4.4}$ & $\scriptstyle{\pm} \textbf{7.9}$ & $\scriptstyle{\pm} \textbf{2.3}$ & $\scriptstyle{\pm} \textbf{3.8}$ \\
       \bottomrule
       \end{tabular}}
        \label{tab4}
    \end{table*}

\begin{table*}[!ht]
\centering
\caption{Performance (DSC, \%) comparison among different sampling methods on each anatomical structure.}\label{tab7}
\resizebox{0.81\textwidth}{!}{%
\begin{tabular}{c|c|c|c|c|c|c|c|c|c|c|c|c|c|c|c|c|c}
\toprule
\textbf{Sampling Method} & \textbf{Sp} & \textbf{RK} & \textbf{LK} & \textbf{GB} & \textbf{Eso} & \textbf{L} & \textbf{St} & \textbf{A} & \textbf{PC} & \textbf{Pan} & \textbf{RAG} & \textbf{LAG} & \textbf{Duo} & \textbf{B} & \textbf{PU} & \textbf{PSV} & \textbf{Average}\\
\midrule
\multirow{2}{*}{Random}  & $94.8$ & $92.7$ & $93.0$ & $75.0$ & $80.9$ & $96.0$ & $90.0$ & $92.0$ & $85.5$ & $83.6$ & $71.7$ & $73.3$ & $72.5$ & $89.5$ & $74.4$ & $73.8$ & $83.7$ \\
                                        & $\scriptstyle{\pm} 6.7$ & $\scriptstyle{\pm} 8.0$ & $\scriptstyle{\pm} 6.2$ & $\scriptstyle{\pm} 23.7$ & $\scriptstyle{\pm} 9.5$ & $\scriptstyle{\pm} 6.6$ & $\scriptstyle{\pm} 12.0$ & $\scriptstyle{\pm} 5.3$ & $\scriptstyle{\pm} 11.0$ & $\scriptstyle{\pm} 9.0$ & $\scriptstyle{\pm} 14.1$ & $\scriptstyle{\pm} 16.0$ & $\scriptstyle{\pm} 17.4$ & $\scriptstyle{\pm} 13.5$ & $\scriptstyle{\pm} 24.6$ & $\scriptstyle{\pm} 15.9$ & $\scriptstyle{\pm} 12.5$ \\
\multirow{2}{*}{MDC} & $95.0$ & $93.6$ & $93.5$ & $77.8$ & $81.3$ & $96.5$ & $91.1$ & $91.3$ & $86.1$ & $84.6$ & $73.7$ & $74.4$ & $73.2$ & $89.0$ & $75.7$ & $74.9$ & $84.5$ \\
                                             & $\scriptstyle{\pm} 7.8$ & $\scriptstyle{\pm} 8.0$ & $\scriptstyle{\pm} 6.7$ & $\scriptstyle{\pm} 23.2$ & $\scriptstyle{\pm} 8.9$ & $\scriptstyle{\pm} 6.4$ & $\scriptstyle{\pm} 11.2$ & $\scriptstyle{\pm} 9.8$ & $\scriptstyle{\pm} 12.1$ & $\scriptstyle{\pm} 9.0$ & $\scriptstyle{\pm} 12.9$ & $\scriptstyle{\pm} 15.1$ & $\scriptstyle{\pm} 18.0$ & $\scriptstyle{\pm} 14.4$ & $\scriptstyle{\pm} 23.1$ & $\scriptstyle{\pm} 15.3$ & $\scriptstyle{\pm} 12.6$ \\
CMD & $\textbf{95.3}$ & $\textbf{93.8}$ & $\textbf{94.0}$ & $\textbf{77.7}$ & $\textbf{82.4}$ & $\textbf{96.5}$ & $\textbf{91.5}$ & $\textbf{92.5}$ & $\textbf{87.0}$ & $\textbf{85.1}$ & $\textbf{75.4}$ & $\textbf{76.5}$ & $\textbf{76.4}$ & $\textbf{90.3}$ & $\textbf{79.2}$ & $\textbf{77.6}$ & $\textbf{85.7}$ \\
                                 (Default) & $\scriptstyle{\pm} \textbf{6.6}$ & $\scriptstyle{\pm} \textbf{7.3}$ & $\scriptstyle{\pm} \textbf{5.2}$ & $\scriptstyle{\pm} \textbf{25.0}$ & $\scriptstyle{\pm} \textbf{9.0}$ & $\scriptstyle{\pm} \textbf{6.4}$ & $\scriptstyle{\pm} \textbf{12.3}$ & $\scriptstyle{\pm} \textbf{6.5}$ & $\scriptstyle{\pm} \textbf{11.2}$ & $\scriptstyle{\pm} \textbf{9.3}$ & $\scriptstyle{\pm} \textbf{12.0}$ & $\scriptstyle{\pm} \textbf{14.6}$ & $\scriptstyle{\pm} \textbf{16.6}$ & $\scriptstyle{\pm} \textbf{13.3}$ & $\scriptstyle{\pm} \textbf{17.6}$ & $\scriptstyle{\pm} \textbf{16.5}$ & $\scriptstyle{\pm} \textbf{11.9}$ \\
\bottomrule
\end{tabular}}
\end{table*}

\begin{figure*}[!h]
  \centering
  \includegraphics[width=0.75\textwidth]{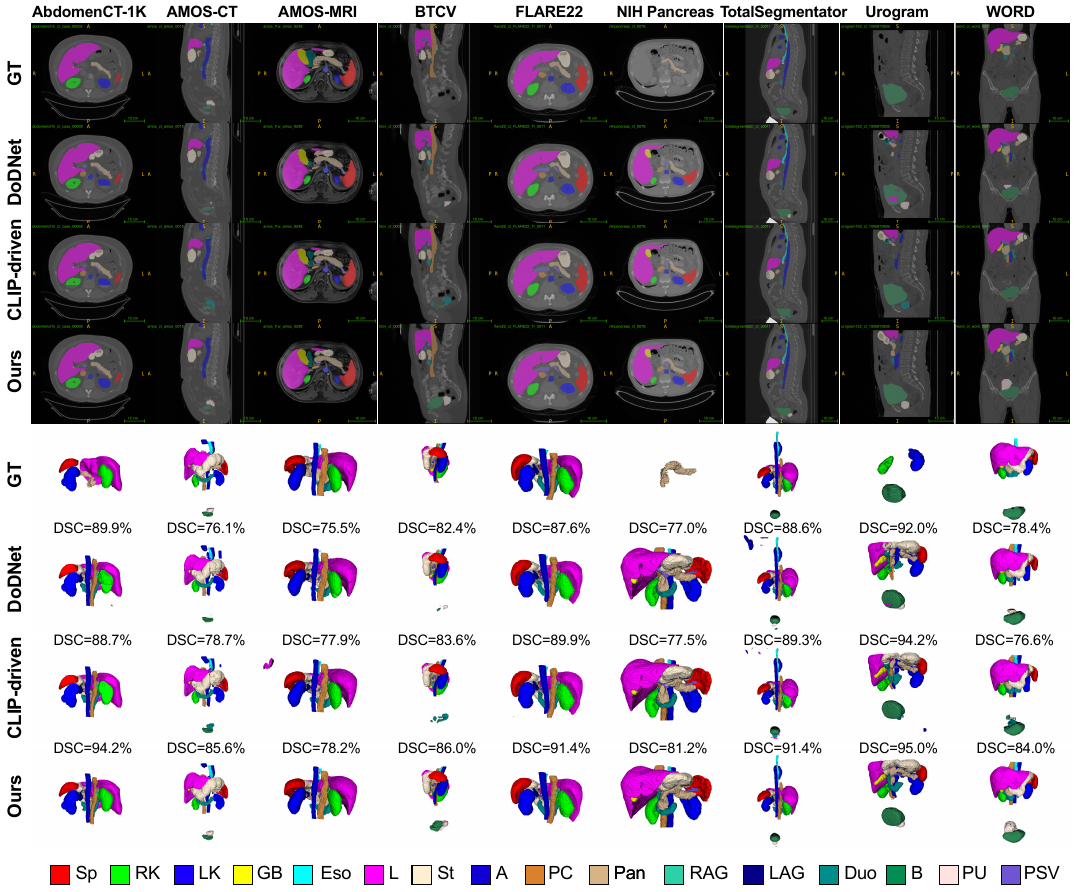}
  \caption{Visual comparison between the ground truth and the predictions generated by DoDNet, CLIP-driven and the proposed method on subjects from different datasets. For a clearer view of detailed differences, zoom in to closely examine the results.}
  \label{fig4}
\end{figure*}

\begin{table*}[!ht]
   \begin{minipage}{0.35\textwidth}
        \centering
        \captionsetup{font=scriptsize}
        \caption{Performance trained with varying numbers of partially labeled images with and without entropy minimization.}
        \resizebox{0.5\textwidth}{!}{%
        \begin{tabular}{c|c}
        \toprule
        \textbf{Setting} & \textbf{DSC} [\%] \\
        \midrule
        3 Sets w/o reg & $82.6 \scriptstyle{\pm} 15.2$ \\
        3 Sets w/ reg &$83.5 \scriptstyle{\pm} 13.9$ \\
        8 Sets w/o reg & $88.2 \scriptstyle{\pm} \textbf{7.0}$ \\
        8 Sets w/ reg & $\textbf{88.7} \scriptstyle{\pm} \textbf{7.0}$ \\
        \bottomrule
        \end{tabular}}
        \label{tab9}
    \end{minipage}%
    \hspace{1.5em}
    \begin{minipage}{0.33\textwidth}
        \centering
        \captionsetup{font=scriptsize}
        \caption{Performance (DSC, \%) on sparsely labeled data with different portions of annotated slices.}
        \resizebox{0.8\textwidth}{!}{%
        \begin{tabular}{c|c|c}
        \toprule
        \textbf{Setting} & \textbf{20\%} & \textbf{100\%} \\
        \midrule
        8 Sets (axial) & $85.1 \scriptstyle{\pm} 11.8$ & $87.8 \scriptstyle{\pm} 8.0$ \\
        8 Sets (sagittal) & $86.2 \scriptstyle{\pm} 9.0$ & $87.8 \scriptstyle{\pm} 7.9$ \\
        8 Sets (coronal) & $86.1 \scriptstyle{\pm} 10.3$ & $87.8 \scriptstyle{\pm} 7.9$ \\
        \bottomrule
        \end{tabular}}
        \label{tab10}
    \end{minipage}%
    \hspace{-0.25em}
    \begin{minipage}{0.31\textwidth}
        \centering
        \captionsetup{font=scriptsize}
        \caption{Performance with mixed training.}
        \resizebox{0.6\textwidth}{!}{%
        \begin{tabular}{c|c}
        \toprule
        \textbf{Setting} & \textbf{DSC} [\%] \\
        \midrule
        8 Sets (axial) & $87.6 \scriptstyle{\pm} 8.3$ \\
        8 Sets (sagittal) & $87.7 \scriptstyle{\pm} 8.5$ \\
        8 Sets (coronal) & $87.6 \scriptstyle{\pm} 7.6$ \\
        \bottomrule
        \end{tabular}}
        \label{tab11}
    \end{minipage}
\end{table*}

\subsection{Results on Partially Labeled Data}
Current methods are limited to utilizing partially labeled data for training. Therefore, we compared our method and state-of-the-art approaches, DoDNet and CLIP-driven, using exclusively partially labeled data. For fair comparisons, we replaced the base network in their original frameworks with the 3D TransUNet and adopted the same hierarchical sampling approach as ours. Notably, we adjusted DoDNet to produce a single-channel output, since DoDNet was originally designed to predict both anatomical structures and associated tumors concurrently whereas our study focused on the former. Furthermore, both DoDNet and CLIP-driven were unable to utilize images lacking annotations, such as those from TotalSegmentator that contain no relevant anatomical structures under study within their ROIs. Consequently, we excluded those images for training DoDNet and CLIP-driven models.

\noindent \textbf{Main results.} Table~\ref{tab2} summarizes the segmentation performance assessed \textit{on a per-subject basis} for DoDNet, CLIP-driven, and the proposed method, which employed different base networks. This evaluation method computed the average DSC of all annotated classes for each testing subject and subsequently averaged them across subjects. During training DoDNet and CLIP-driven, we observed that they exhibited significantly slower convergence rates and thus doubled the training time compared to our proposed method. Our experimental results indicated that, using the same 3D TransUNet as the base network, our approach achieved an impressive average DSC of $88.7\%$ on the testing set, surpassing both DoDNet and CLIP-driven, which attained average DSCs of $83.5\%$ and $83.3\%$, respectively.

Further insights into performance were gleaned by examining segmentation performance \textit{on each anatomical structure}, as depicted in Table~\ref{tab3}. This analysis involved averaging DSCs across individual images with specific structures annotated, revealing the superior segmentation performance of the proposed method over DoDNet and CLIP-driven. Remarkably, our proposed method, employing 3D TransUNet as the base network, achieved an average DSC of $85.7\%$ for individual structures, outperforming DoDNet and CLIP-driven by $5.7\%$ and $5.0\%$, respectively.

Moreover, we conducted a comparative evaluation of their performance and undertook a visual comparison across each dataset, as illustrated in Table~\ref{tab4} and Fig.~\ref{fig4}, which accentuated the consistently superior performance of the proposed method across all datasets.

To evaluate model generalizability to unseen datasets, we additionally trained a model using only AMOS, BTCV, and FLARE22 as training data. As summarized in Table~\ref{tab8}, this model achieved an average DSC of $83.5\%$ on the testing set, lower than the one trained with all data, which was expected. Notably, using all eight datasets for training yielded a model with superior prostate/uterus segmentation performance compared to the one trained using only three datasets, despite the additional datasets lacking annotations for the prostate/uterus. This highlights the advantages of fine-grained segmentation.

Additionally, it should be noted that both DoDNet and CLIP-driven demand $16$ forward passes to predict the desired anatomical structures, whereas our proposed method achieved the same with just a single forward pass, indicating our method's substantially improved efficiency.

\noindent \textbf{Effect of base network.} We compared 3D TransUNet, our custom design, and the default network with other top performers in medical image segmentation, including Unet\text{++} \cite{zhou2019unet++}, Swin UNETR \cite{hatamizadeh2021swin}, and MedNeXt-L \cite{roy2023mednext}. All training configurations were consistent across different networks, with the patch size for Swin UNETR adjusted to $96\times96\times96$ due to memory constraints. Results in Table~\ref{tab2} demonstrated that while MedNeXt achieved comparable performance to 3D TransUNet on a per-subject basis, Unet\text{++} and Swin UNETR exhibited inferior performance by over $1\%$ in terms of DSC. A detailed class-wise comparison in Table~\ref{tab3} highlighted 3D TransUNet's stronger performance compared to all others, including MedNeXt. Notably, our approach was largely not affected by the choices of its base network.

\noindent \textbf{Effect of sampling method.} To evaluate the benefits of hierarchical sampling, we compared three strategies: 1) sampling following the class$\rightarrow$modality$\rightarrow$dataset hierarchy (CMD, the default), 2) sampling following the modality$\rightarrow$dataset$\rightarrow$class hierarchy (MDC), and 3) random sampling that selects an image randomly from the dataset and then chooses a random location within the image volume as the center for training patches. The results presented in Table~\ref{tab6} indicated that our approach outperformed random sampling by $1.2\%$ in overall DSC. While MDC exhibited a $0.4\%$ improvement in DSC over Random, it lagged $0.8\%$ behind CMD. These advantages were particularly noticeable in the comparison on each anatomical structure, especially for smaller structures, such as LAG and RAG, as emphasized in Table~\ref{tab7}.

\noindent \textbf{Effect of regularization term.} The comparison between models with and without the entropy minimization regularization term is outlined in Table~\ref{tab9}. Despite a decrease in performance gain with larger training datasets, consistent enhancements were observed across various dataset sizes. When all $8$ datasets were used for training, the addition of the regularization term resulted in a DSC improvement of $0.5\%$. Notably, when training the model with $3$ datasets, including AMOS, BTCV, and FALRE22, a larger improvement of $0.9\%$ in terms of DSC was achieved.

\subsection{Results on Sparsely Labeled Data}
Our method stands out from existing ones in its capability of handling sparsely labeled data, a critical feature that enhances its applicability in real-world scenarios where the annotation budget is limited and/or data are sparsely labeled. For demonstration purpose, we conducted experiments in which we selectively chose slices from axial, sagittal, or coronal views for training. The experimental results, summarized in Table~\ref{tab10}, revealed that models trained with only $20\%$ of slices (evenly spaced) achieved an impressive average DSC ranging from $85.1\%$ to $86.2\%$, outperforming baseline methods trained with all slices (cf. Table~\ref{tab2}). For comparison, we trained three additional models using the same data as in the partially labeled experiments (i.e., trained with $100\%$ slices), but with slice-by-slice loss calculation to simulate sparsely labeled data conditions. These results served as an upper bound and demonstrated the consistent performance of our method across different views.

\subsection{Results on Hybrid Data}
We conducted experiments using both partially and sparsely labeled data to mimic real-world scenarios. Our mixed training approach utilized AMOS, BTCV, and FLARE22 entirely, and $20\%$ of evenly spaced slices of the other five datasets from axial, sagittal, or coronal views, respectively. Table~\ref{tab11} demonstrates that this hybrid data approach achieved DSCs of $87.6\%$, $87.7\%$, and $87.6\%$ for the respective models. In contrast, the model trained solely with $3$ partially labeled datasets attained an average DSC of $83.5\%$ (cf. Table~\ref{tab8}). Integrating sparsely labeled data notably improved the performance by approximately $4.1\%$.

\section{Conclusions}
We have developed a novel weakly-supervised medical image segmentation approach that effectively utilizes multi-source partially and sparsely labeled data for training. Our method addresses data limitations of large, diverse, fully annotated datasets, enhancing label efficiency and reducing annotation efforts through the utilization of weakly annotated data. By integrating strategies for model self-disambiguation, prior knowledge incorporation, and imbalance mitigation, our approach establishes a solid foundation for training versatile and reliable segmentation models.


{
    \small
    \bibliographystyle{ieeenat_fullname}
    \bibliography{main}
}

\clearpage
\setcounter{page}{1}
\maketitlesupplementary

\appendix

\renewcommand{\thesection}{\arabic{section}}
\section{More Details about Datasets}
Details of the seven public datasets are provided in their corresponding papers. Regarding the private dataset, it comprises $122$ contrast-enhanced CT images from patients undergoing urinary system examinations. The images have a uniform matrix size of $512 \times 512$, with a variable number of 2D slices ranging from $62$ to $685$. Pixel spacing ranges from $0.607$ to $0.977$ $\text{mm}$, and slice thickness varies from $1.0$ to $3.0$ $\text{mm}$. Urologists annotated the kidney, bladder, and ureters in each image. For this study, only the masks of the two kidneys and the bladder were retained.

\section{More Details about Network Architecture}
Table~\ref{network_architecture} presents the architecture for 3D TransUNet. The structure of 3D TransUNet is asymmetric, featuring a greater number of layers in the encoder compared to the decoder. Both the encoder and decoder are composed of $5$ stages, wherein spatial sizes progressively decrease by $50\%$ from stage $1$ to stage $5$ in a sequential manner.

The network's building blocks are shown in brackets. All blocks, except those in stage 5, comprise two consecutive convolutional layers. The adjacent pair of numbers within each bracket represent the input channels and output channels of a convolutional layer. A skip connection \cite{he2016deep} is added when the input channels of the first convolutional layer is different from the output channels of the second convolutional layer within each building block in stages 1--4. In accordance with \cite{chen2021transunet}, we employed weight normalization \cite{salimans2016weight} in every convolutional layer to expedite training. Subsequent to each convolution operation, instance normalization \cite{ulyanov2016instance} and rectified linear unit activation are applied. Downsampling and upsampling are executed through trilinear interpolation.

At the bottleneck, four multi-head attention layers were incorporated, each with eight heads. The size of each attention head for query, key, and value was set to be $512$.

\begin{table}[!ht]
        \centering
        \caption{Network architecture.}
        \resizebox{0.8\columnwidth}{!}{%
        \begin{tabular}{c|c|c}
            \toprule
             & \textbf{Encoder} & \textbf{Decoder} \\
            \midrule
            {Stage 1} & \{1, 32, 64\} & \{128, 64, 64\} \\
            \midrule
            \multirow{2}{*}{Stage 2} & \{64, 64, 128\} & \{128, 64, 64\} \\
                                                 & \{128, 128, 256\} & \\
            \midrule
            \multirow{3}{*}{Stage 3} & \{256, 128, 256\} & \{512, 64, 64\}  \\ 
                                                 & \{256, 128, 256\} & \\
                                                 & \{256, 128, 512\} & \\
            
            \midrule
            \multirow{4}{*}{Stage 4} & \{512, 256, 512\} & \{1024, 256, 256\} \\
                                                  & \{512, 256, 512\} & \\
                                                  & \{512, 256, 512\} & \\
                                                  & \{512, 256, 1024\} & \\
            \midrule
            {Stage 5} & \{1024, 512\} & \{512, 512\} \\
            \bottomrule
         \end{tabular}}
        \label{network_architecture}
\end{table}

\section{More Results on Partially Labeled Data}
\noindent \textbf{Effect of patch size.} We conducted experiments with two different patch sizes, namely $96 \times 96 \times 96$ and $112 \times 112 \times 112$. Larger patch sizes were not explored due to limitations in GPU memory. As indicated in Table~\ref{effect_of_patch_size}, employing a patch size of $96 \times 96 \times 96$ resulted in an average DSC of $87.9\%$, which is $0.8\%$ DSC lower than the performance achieved with a patch size of $112 \times 112 \times 112$. These findings underscore the advantageous impact of using a larger patch for abdominal organ segmentation, since increased patch size contributes to a more comprehensive inclusion of contextual information.

\begin{table}[!ht]
        \centering
        \caption{Performance trained with different patch sizes.}
        \begin{tabular}{c|c}
        \toprule
        \textbf{Patch Size} & \textbf{DSC} [\%] \\
        \midrule
        $96 \times 96 \times 96$ & $87.9  \scriptstyle{\pm} 8.1$ \\
        $112 \times 112 \times 112$ & $\textbf{88.7} \scriptstyle{\pm} \textbf{7.0}$ \\
        \bottomrule
        \end{tabular}
        \label{effect_of_patch_size}
\end{table}

\noindent \textbf{Effect of voxel spacing.} In our experiments, we standardized the voxel spacing for all images to facilitate model training. To assess the influence of voxel spacing on model performance, we conducted experiments with three different voxel spacings: $1.5 \times 1.5 \times 1.5$ $\text{mm}^{3}$, $2.0 \times 2.0 \times 2.0$ $\text{mm}^{3}$, and $2.5 \times 2.5 \times 2.5$ $\text{mm}^{3}$. As indicated in Table~\ref{effect_of_voxel_spacing}, employing a voxel spacing of $1.5 \times 1.5 \times 1.5$ $\text{mm}^{3}$ and $2.5 \times 2.5 \times 2.5$ $\text{mm}^{3}$ led to a performance decrease of $0.3\%$ and $0.6\%$ in terms of average DSC, respectively. The diminished performance with a voxel spacing of $1.5 \times 1.5 \times 1.5$ $\text{mm}^{3}$ can be attributed to reduced contextual information within the input image patch. Conversely, the inferior performance with a voxel spacing of $2.5 \times 2.5 \times 2.5$ $\text{mm}^{3}$ is likely due to information loss during downsampling, particularly impacting small structure segmentation.

\begin{table}[!ht]
        \centering
        \caption{Performance trained with different patch sizes.}
        \begin{tabular}{c|c}
        \toprule
        \textbf{Voxel Size} & \textbf{DSC} [\%] \\
        \midrule
        $1.5 \times 1.5 \times 1.5$ & $88.4  \scriptstyle{\pm} 7.7$ \\
        $2.0 \times 2.0 \times 2.0$ & $\textbf{88.7} \scriptstyle{\pm} \textbf{7.0}$ \\
        $2.5 \times 2.5 \times 2.5$ & $88.1  \scriptstyle{\pm} 7.8$ \\
        \bottomrule
        \end{tabular}
        \label{effect_of_voxel_spacing}
\end{table}


\section{More Results on Sparsely Labeled Data}
Tables~\ref{supp_tab1} and~\ref{supp_tab2} provide a detailed comparison of the performance for each anatomical structure and across each dataset, respectively. Visual results of a randomly selected subject from each dataset are presented in the second and third columns of Fig.~\ref{supp_fig1}. These results align with the findings in Table 8, highlighting the consistent success of our method across different views. Notably, even with the utilization of only $20\%$ of incompletely annotated slices for training, our method demonstrates commendable performance across the structures of interest and datasets.

\section{More Results on Hybrid Data}
\begin{figure*}[!h]
  \centering
  \includegraphics[width=0.67\textwidth]{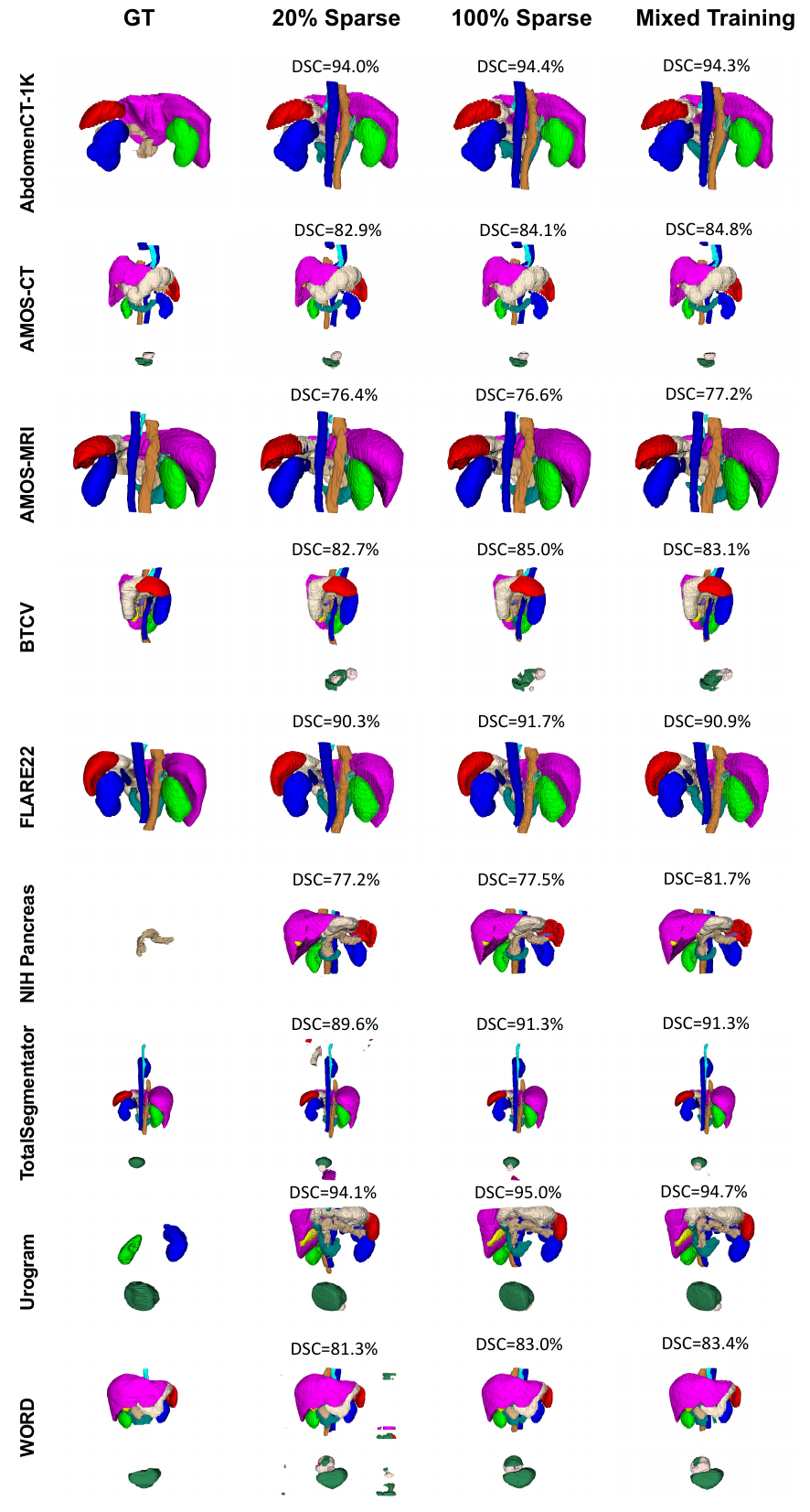}
  \caption{Visual comparisons between the ground truth and predictions from models trained with 20\% slices of the axial view, 100\% slices of the axial view (loss is computed slice-wise to emulate sparsely labeled data), and hybrid data (the entirety of AMOS, BTCV, and FLARE22 is utilized, while 20\% slices of the axial view are taken from other datasets for training) on subjects from various datasets. For a clearer view of detailed differences, zoom in to closely examine the results.}
  \label{supp_fig1}
\end{figure*}

Tables~\ref{supp_tab3} and~\ref{supp_tab4} present a comprehensive comparison of performance for each anatomical structure and across each dataset, respectively. Visual results of a randomly selected subject from each dataset are displayed in the fourth column of Fig.~\ref{supp_fig1}. These results concur with the findings in Table 9, underscoring the effectiveness of our method in utilizing a mixture of partially and sparsely labeled data for model training. 

\begin{table*}[!ht]
\centering
\caption{Performance (DSC, \%) comparison on each anatomical structure using different portions of annotated slices.}\label{supp_tab1}
\resizebox{0.96\textwidth}{!}{%
\begin{tabular}{c|c|c|c|c|c|c|c|c|c|c|c|c|c|c|c|c|c|c}
\toprule
\textbf{Setting} & \textbf{View} & \textbf{Sp} & \textbf{RK} & \textbf{LK} & \textbf{GB} & \textbf{Eso} & \textbf{L} & \textbf{St} & \textbf{A} & \textbf{PC} & \textbf{Pan} & \textbf{RAG} & \textbf{LAG} & \textbf{Duo} & \textbf{B} & \textbf{PU} & \textbf{PSV} & \textbf{Average}\\
\midrule
\multirow{6}{*}{\centering $20\%$} & \multirow{2}{*}{\centering Axial} & $93.5$ & $92.6$ & $92.4$ & $74.3$ & $78.8$ & $95.4$ & $89.5$ & $89.4$ & $83.2$ & $83.1$ & $70.8$ & $72.4$ & $70.5$ & $85.7$ & $71.1$ & $70.7$ & $82.0$ \\
                                  & & $\scriptstyle{\pm} 9.3$ & $\scriptstyle{\pm} 10.9$ & $\scriptstyle{\pm} 9.9$ & $\scriptstyle{\pm} 26.4$ & $\scriptstyle{\pm} 12.0$ & $\scriptstyle{\pm} 9.6$ & $\scriptstyle{\pm} 13.7$ & $\scriptstyle{\pm} 12.1$ & $\scriptstyle{\pm} 15.0$ & $\scriptstyle{\pm} 11.4$ & $\scriptstyle{\pm} 14.1$ & $\scriptstyle{\pm} 16.3$ & $\scriptstyle{\pm} 23.5$ & $\scriptstyle{\pm} 17.2$ & $\scriptstyle{\pm} 22.9$ & $\scriptstyle{\pm} 20.1$ & $\scriptstyle{\pm} 15.3$ \\
 & \multirow{2}{*}{\centering Sagittal} & $94.5$ & $92.5$ & $91.9$ & $75.6$ & $76.9$ & $96.1$ & $90.5$ & $91.0$ & $85.0$ & $83.5$ & $71.6$ & $73.1$ & $71.7$ & $88.6$ & $74.2$ & $72.3$ & $83.1$ \\
                                  & & $\scriptstyle{\pm} 8.1$ & $\scriptstyle{\pm} 11.1$ & $\scriptstyle{\pm} 9.0$ & $\scriptstyle{\pm} 25.5$ & $\scriptstyle{\pm} 11.3$ & $\scriptstyle{\pm} 7.1$ & $\scriptstyle{\pm} 12.2$ & $\scriptstyle{\pm} 8.7$ & $\scriptstyle{\pm} 12.9$ & $\scriptstyle{\pm} 9.3$ & $\scriptstyle{\pm} 13.0$ & $\scriptstyle{\pm} 14.6$ & $\scriptstyle{\pm} 19.7$ & $\scriptstyle{\pm} 14.6$ & $\scriptstyle{\pm} 24.5$ & $\scriptstyle{\pm} 18.5$ & $\scriptstyle{\pm} 13.8$ \\
 & \multirow{2}{*}{\centering Coronal} & $94.8$ & $92.6$ & $92.7$ & $75.1$ & $76.5$ & $96.0$ & $90.7$ & $90.6$ & $85.5$ & $83.5$ & $71.3$ & $71.6$ & $73.3$ & $88.2$ & $73.3$ & $74.3$ & $83.1$ \\
                                  & & $\scriptstyle{\pm} 6.8$ & $\scriptstyle{\pm} 10.8$ & $\scriptstyle{\pm} 8.0$ & $\scriptstyle{\pm} 24.3$ & $\scriptstyle{\pm} 12.6$ & $\scriptstyle{\pm} 7.9$ & $\scriptstyle{\pm} 12.0$ & $\scriptstyle{\pm} 7.8$ & $\scriptstyle{\pm} 13.5$ & $\scriptstyle{\pm} 10.5$ & $\scriptstyle{\pm} 14.6$ & $\scriptstyle{\pm} 16.6$ & $\scriptstyle{\pm} 18.6$ & $\scriptstyle{\pm} 15.3$ & $\scriptstyle{\pm} 20.2$ & $\scriptstyle{\pm} 17.6$ & $\scriptstyle{\pm} 13.6$ \\

\midrule
\multirow{6}{*}{\centering $100\%$} & \multirow{2}{*}{\centering Axial} & $95.0$ & $93.5$ & $93.9$ & $76.1$ & $80.7$ & $96.6$ & $91.3$ & $91.7$ & $86.2$ & $84.6$ & $74.5$ & $74.9$ & $75.0$ & $89.2$ & $76.2$ & $76.8$ & $84.8$ \\
                                  & & $\scriptstyle{\pm} 7.0$ & $\scriptstyle{\pm} 8.4$ & $\scriptstyle{\pm} 5.8$ & $\scriptstyle{\pm} 26.1$ & $\scriptstyle{\pm} 10.2$ & $\scriptstyle{\pm} 5.4$ & $\scriptstyle{\pm} 11.5$ & $\scriptstyle{\pm} 9.0$ & $\scriptstyle{\pm} 12.3$ & $\scriptstyle{\pm} 8.8$ & $\scriptstyle{\pm} 12.0$ & $\scriptstyle{\pm} 14.3$ & $\scriptstyle{\pm} 18.3$ & $\scriptstyle{\pm} 13.6$ & $\scriptstyle{\pm} 20.1$ & $\scriptstyle{\pm} 15.8$ & $\scriptstyle{\pm} 12.4$ \\
 & \multirow{2}{*}{\centering Sagittal} & $94.6$ & $93.1$ & $93.7$ & $76.1$ & $80.4$ & $96.2$ & $91.2$ & $91.1$ & $86.4$ & $84.6$ & $72.7$ & $73.9$ & $73.9$ & $89.0$ & $76.9$ & $75.8$ & $84.3$ \\
                                  & & $\scriptstyle{\pm} 8.6$ & $\scriptstyle{\pm} 8.7$ & $\scriptstyle{\pm} 5.1$ & $\scriptstyle{\pm} 25.1$ & $\scriptstyle{\pm} 8.6$ & $\scriptstyle{\pm} 7.1$ & $\scriptstyle{\pm} 12.0$ & $\scriptstyle{\pm} 8.8$ & $\scriptstyle{\pm} 10.4$ & $\scriptstyle{\pm} 9.9$ & $\scriptstyle{\pm} 14.2$ & $\scriptstyle{\pm} 15.2$ & $\scriptstyle{\pm} 18.2$ & $\scriptstyle{\pm} 14.4$ & $\scriptstyle{\pm} 23.8$ & $\scriptstyle{\pm} 17.8$ & $\scriptstyle{\pm} 13.0$ \\
 & \multirow{2}{*}{\centering Coronal} & $95.2$ & $93.5$ & $93.7$ & $77.7$ & $81.3$ & $96.5$ & $91.1$ & $91.7$ & $86.4$ & $84.4$ & $73.0$ & $74.6$ & $74.8$ & $89.6$ & $73.7$ & $75.2$ & $84.5$ \\
                                  & & $\scriptstyle{\pm} 6.6$ & $\scriptstyle{\pm} 7.6$ & $\scriptstyle{\pm} 5.6$ & $\scriptstyle{\pm} 23.5$ & $\scriptstyle{\pm} 8.8$ & $\scriptstyle{\pm} 6.4$ & $\scriptstyle{\pm} 12.2$ & $\scriptstyle{\pm} 9.1$ & $\scriptstyle{\pm} 12.0$ & $\scriptstyle{\pm} 9.7$ & $\scriptstyle{\pm} 13.6$ & $\scriptstyle{\pm} 14.7$ & $\scriptstyle{\pm} 17.1$ & $\scriptstyle{\pm} 12.8$ & $\scriptstyle{\pm} 19.1$ & $\scriptstyle{\pm} 16.6$ & $\scriptstyle{\pm} 12.2$ \\                                                                
\bottomrule
\end{tabular}}
\end{table*}

\begin{table*}[!ht]
        \centering
        \caption{Performance (DSC, \%) comparison on each dataset using different portions of annotated slices.}\label{supp_tab2}
        \resizebox{0.96\textwidth}{!}{%
        \begin{tabular}{c|c|c|c|c|c|c|c|c|c|c}
        \toprule
        \textbf{Setting} & \textbf{View} & \textbf{AbCT-1K} & \textbf{AMOS-CT} & \textbf{AMOS-MRI} & \textbf{BTCV} & \textbf{FLARE22} & \textbf{NIH-Pan} & \textbf{TotalSeg} & \textbf{Urogram} & \textbf{WORD} \\
        \midrule
        \multirow{6}{*}{$20\%$} & \multirow{2}{*}{Axial} & $92.7$ & $82.7$ & $80.0$ & $76.3$ & $89.8$ & $83.5$ & $79.3$ & $92.7$ & $82.0$ \\
                                       & & $\scriptstyle{\pm} 2.0$ & $\scriptstyle{\pm} 8.1$ & $\scriptstyle{\pm} 10.1$ & $\scriptstyle{\pm} 5.8$ & $\scriptstyle{\pm} 1.5$ & $\scriptstyle{\pm} 4.7$ & $\scriptstyle{\pm} 15.2$ & $\scriptstyle{\pm} 2.6$ & $\scriptstyle{\pm} 4.4$ \\
         & \multirow{2}{*}{Sagittal} & $92.7$ & $83.4$ & $80.5$ & $76.5$ & $89.4$ & $83.2$ & $82.0$ & $92.3$ & $82.7$ \\
                                       & & $\scriptstyle{\pm} 1.9$ & $\scriptstyle{\pm} 7.6$ & $\scriptstyle{\pm} 8.5$ & $\scriptstyle{\pm} 6.5$ & $\scriptstyle{\pm} 1.5$ & $\scriptstyle{\pm} 5.3$ & $\scriptstyle{\pm} 11.3$ & $\scriptstyle{\pm} 3.0$ & $\scriptstyle{\pm} 4.4$ \\
         & \multirow{2}{*}{Coronal} & $92.8$ & $83.5$ & $79.2$ & $76.1$ & $89.8$ & $83.3$ & $81.6$ & $92.6$ & $82.5$ \\
                                       & & $\scriptstyle{\pm} 2.3$ & $\scriptstyle{\pm} 7.2$ & $\scriptstyle{\pm} 9.8$ & $\scriptstyle{\pm} 7.8$ & $\scriptstyle{\pm} 1.6$ & $\scriptstyle{\pm} 5.8$ & $\scriptstyle{\pm} 13.1$ & $\scriptstyle{\pm} 3.0$ & $\scriptstyle{\pm} 4.2$ \\
      
        \midrule
        \multirow{6}{*}{$100\%$} & \multirow{2}{*}{Axial} & $93.4$ & $85.1$ & $80.9$ & $77.9$ & $90.5$ & $83.8$ & $84.7$ & $93.0$ & $83.7$ \\
                                       & & $\scriptstyle{\pm} 1.7$ & $\scriptstyle{\pm} 6.3$ & $\scriptstyle{\pm} 9.0$ & $\scriptstyle{\pm} 6.4$ & $\scriptstyle{\pm} 1.8$ & $\scriptstyle{\pm} 5.4$ & $\scriptstyle{\pm} 9.7$ & $\scriptstyle{\pm} 3.0$ & $\scriptstyle{\pm} 4.1$ \\
         & \multirow{2}{*}{Sagittal} & $93.0$ & $84.2$ & $81.5$ & $78.7$ & $90.4$ & $84.5$ & $84.2$ & $93.0$ & $83.5$ \\
                                       & & $\scriptstyle{\pm} 2.2$ & $\scriptstyle{\pm} 7.7$ & $\scriptstyle{\pm} 7.7$ & $\scriptstyle{\pm} 5.5$ & $\scriptstyle{\pm} 1.0$ & $\scriptstyle{\pm} 5.2$ & $\scriptstyle{\pm} 9.8$ & $\scriptstyle{\pm} 2.9$ & $\scriptstyle{\pm} 4.4$ \\
         & \multirow{2}{*}{Coronal} & $93.3$ & $84.5$ & $81.1$ & $78.1$ & $90.3$ & $84.4$ & $85.0$ & $93.0$ & $83.7$ \\
                                       & & $\scriptstyle{\pm} 2.0$ & $\scriptstyle{\pm} 6.9$ & $\scriptstyle{\pm} 8.3$ & $\scriptstyle{\pm} 5.7$ & $\scriptstyle{\pm} 1.3$ & $\scriptstyle{\pm} 4.2$ & $\scriptstyle{\pm} 9.5$ & $\scriptstyle{\pm} 3.2$ & $\scriptstyle{\pm} 4.0$ \\
                                                
       \bottomrule
       \end{tabular}}
    \end{table*}

\begin{table*}[!ht]
\centering
\caption{Performance (DSC, \%) comparison on each anatomical structure using mixed training.}\label{supp_tab3}
\resizebox{0.96\textwidth}{!}{%
\begin{tabular}{c|c|c|c|c|c|c|c|c|c|c|c|c|c|c|c|c|c}
\toprule
\textbf{View} & \textbf{Sp} & \textbf{RK} & \textbf{LK} & \textbf{GB} & \textbf{Eso} & \textbf{L} & \textbf{St} & \textbf{A} & \textbf{PC} & \textbf{Pan} & \textbf{RAG} & \textbf{LAG} & \textbf{Duo} & \textbf{B} & \textbf{PU} & \textbf{PSV} & \textbf{Average}\\
\midrule
\multirow{2}{*}{\centering Axial} & $95.1$ & $93.4$ & $93.8$ & $75.8$ & $81.6$ & $96.4$ & $90.7$ & $91.7$ & $86.1$ & $84.5$ & $74.6$ & $75.7$ & $73.4$ & $89.0$ & $76.6$ & $73.2$ & $84.5$ \\
                                  & $\scriptstyle{\pm} 9.3$ & $\scriptstyle{\pm} 10.9$ & $\scriptstyle{\pm} 9.9$ & $\scriptstyle{\pm} 26.4$ & $\scriptstyle{\pm} 12.0$ & $\scriptstyle{\pm} 9.6$ & $\scriptstyle{\pm} 13.7$ & $\scriptstyle{\pm} 12.1$ & $\scriptstyle{\pm} 15.0$ & $\scriptstyle{\pm} 11.4$ & $\scriptstyle{\pm} 14.1$ & $\scriptstyle{\pm} 16.3$ & $\scriptstyle{\pm} 23.5$ & $\scriptstyle{\pm} 17.2$ & $\scriptstyle{\pm} 22.9$ & $\scriptstyle{\pm} 20.1$ & $\scriptstyle{\pm} 15.3$ \\
\multirow{2}{*}{\centering Sagittal} & $95.1$ & $93.2$ & $93.7$ & $77.6$ & $80.6$ & $96.6$ & $91.6$ & $92.1$ & $86.5$ & $84.9$ & $74.1$ & $75.5$ & $75.3$ & $89.2$ & $76.6$ & $75.9$ & $84.9$ \\
                                  & $\scriptstyle{\pm} 7.1$ & $\scriptstyle{\pm} 10.3$ & $\scriptstyle{\pm} 6.6$ & $\scriptstyle{\pm} 24.0$ & $\scriptstyle{\pm} 10.7$ & $\scriptstyle{\pm} 6.5$ & $\scriptstyle{\pm} 11.5$ & $\scriptstyle{\pm} 7.1$ & $\scriptstyle{\pm} 11.2$ & $\scriptstyle{\pm} 8.5$ & $\scriptstyle{\pm} 13.7$ & $\scriptstyle{\pm} 14.3$ & $\scriptstyle{\pm} 17.5$ & $\scriptstyle{\pm} 14.7$ & $\scriptstyle{\pm} 26.7$ & $\scriptstyle{\pm} 17.4$ & $\scriptstyle{\pm} 13.0$ \\
\multirow{2}{*}{\centering Coronal} & $95.1$ & $93.8$ & $93.8$ & $77.2$ & $80.6$ & $96.4$ & $91.3$ & $91.9$ & $86.3$ & $84.7$ & $73.9$ & $75.7$ & $75.1$ & $89.5$ & $77.6$ & $76.7$ & $85.0$ \\
                                  & $\scriptstyle{\pm} 6.8$ & $\scriptstyle{\pm} 6.4$ & $\scriptstyle{\pm} 6.0$ & $\scriptstyle{\pm} 25.0$ & $\scriptstyle{\pm} 11.3$ & $\scriptstyle{\pm} 7.3$ & $\scriptstyle{\pm} 12.4$ & $\scriptstyle{\pm} 7.8$ & $\scriptstyle{\pm} 12.4$ & $\scriptstyle{\pm} 8.9$ & $\scriptstyle{\pm} 14.2$ & $\scriptstyle{\pm} 13.8$ & $\scriptstyle{\pm} 18.2$ & $\scriptstyle{\pm} 14.2$ & $\scriptstyle{\pm} 23.6$ & $\scriptstyle{\pm} 17.4$ & $\scriptstyle{\pm} 12.9$ \\
\bottomrule
\end{tabular}}
\end{table*}

\begin{table*}[!ht]
        \centering
        \caption{Performance (DSC, \%) comparison on each dataset using mixed training.}\label{supp_tab4}
        \resizebox{0.96\textwidth}{!}{%
        \begin{tabular}{c|c|c|c|c|c|c|c|c|c}
        \toprule
        \textbf{View} & \textbf{AbCT-1K} & \textbf{AMOS-CT} & \textbf{AMOS-MRI} & \textbf{BTCV} & \textbf{FLARE22} & \textbf{NIH-Pan} & \textbf{TotalSeg} & \textbf{Urogram} & \textbf{WORD} \\
        \midrule
        \multirow{2}{*}{Axial} & $93.5$ & $85.1$ & $80.9$ & $76.5$ & $90.8$ & $84.3$ & $84.3$ & $93.3$ & $83.2$ \\
                                       & $\scriptstyle{\pm} 1.9$ & $\scriptstyle{\pm} 6.5$ & $\scriptstyle{\pm} 8.9$ & $\scriptstyle{\pm} 5.7$ & $\scriptstyle{\pm} 1.0$ & $\scriptstyle{\pm} 5.2$ & $\scriptstyle{\pm} 10.0$ & $\scriptstyle{\pm} 3.0$ & $\scriptstyle{\pm} 4.2$ \\
        \multirow{2}{*}{Sagittal} & $93.3$ & $85.1$ & $80.7$ & $77.1$ & $90.8$ & $84.2$ & $84.7$ & $93.1$ & $83.5$ \\
                                       & $\scriptstyle{\pm} 1.8$ & $\scriptstyle{\pm} 7.4$ & $\scriptstyle{\pm} 8.5$ & $\scriptstyle{\pm} 7.2$ & $\scriptstyle{\pm} 0.9$ & $\scriptstyle{\pm} 4.4$ & $\scriptstyle{\pm} 10.6$ & $\scriptstyle{\pm} 3.1$ & $\scriptstyle{\pm} 4.0$ \\
        \multirow{2}{*}{Coronal} & $93.3$ & $85.1$ & $80.8$ & $77.6$ & $90.5$ & $84.5$ & $84.1$ & $93.3$ & $83.6$ \\
                                       & $\scriptstyle{\pm} 1.7$ & $\scriptstyle{\pm} 7.2$ & $\scriptstyle{\pm} 9.3$ & $\scriptstyle{\pm} 6.6$ & $\scriptstyle{\pm} 1.1$ & $\scriptstyle{\pm} 4.8$ & $\scriptstyle{\pm} 12.7$ & $\scriptstyle{\pm} 2.7$ & $\scriptstyle{\pm} 4.1$ \\
       \bottomrule
       \end{tabular}}
    \end{table*}

\end{document}